%% file: main.tex
\pdfoutput=1
\documentclass{article}

\usepackage{microtype}
\usepackage{graphicx}
\usepackage{booktabs}
\usepackage{hyperref}

\usepackage{preprint}
\hypersetup{pdftitle={Symmetry Breaking Is the Price of Finite Capacity},
  pdfauthor={Maggie Chlon and Leon Chlon},
  pdfkeywords={symmetry breaking, equivariance, attention, low rank, representation theory},
  colorlinks=true,linkcolor=blue!50!black,citecolor=blue!50!black,urlcolor=blue!50!black}

\usepackage{amsmath,amssymb,mathtools,amsthm}
\usepackage[capitalize,noabbrev]{cleveref}
\usepackage[skins,breakable]{tcolorbox}
\usepackage{enumitem}
\usepackage{array}
\usepackage{xcolor}

\theoremstyle{plain}
\newtheorem{theorem}{Theorem}[section]
\newtheorem{proposition}[theorem]{Proposition}
\newtheorem{lemma}[theorem]{Lemma}
\newtheorem{corollary}[theorem]{Corollary}
\theoremstyle{definition}

\newtheorem{example}[theorem]{Example}
\theoremstyle{remark}
\newtheorem{remark}[theorem]{Remark}

\newtcolorbox{methodbox}[1]{enhanced,breakable,colback=blue!3,colframe=blue!40!black,boxrule=0.5pt,arc=1.5pt,
  left=4pt,right=4pt,top=3pt,bottom=3pt,title={#1},fonttitle=\bfseries\small,fontupper=\small,
  colbacktitle=blue!12,coltitle=black}

\newcommand{\R}{\mathbb{R}}
\newcommand{\E}{\mathbb{E}}
\newcommand{\Sym}{\mathrm{Sym}}
\newcommand{\tr}{\operatorname{tr}}
\newcommand{\rank}{\operatorname{rank}}
\newcommand{\Stab}{\operatorname{Stab}}

\newcommand{\F}{\mathcal{F}}
\newcommand{\Hd}{\mathcal{H}}
\newcommand{\X}{\mathcal{X}}
\newcommand{\inner}[2]{\langle #1,#2\rangle}
\setlist{itemsep=1pt,topsep=2pt,leftmargin=1.2em}

\icmltitlerunning{Symmetry Breaking Is the Price of Finite Capacity}

\begin{document}

\twocolumn[
\icmltitle{Symmetry Breaking Is the Price of Finite Capacity}
\icmlsetsymbol{equal}{*}
\begin{icmlauthorlist}
\icmlauthor{Maggie Chlon}{hassana}
\icmlauthor{Leon Chlon}{hassana}
\end{icmlauthorlist}
\icmlaffiliation{hassana}{Hassana Labs}
\icmlkeywords{symmetry breaking, equivariance, attention, low rank, representation theory}
\vskip 0.3in
]

\printAffiliationsAndNotice{}

\begin{abstract}
If rotating a problem never changes its answer, one might expect its best model to be rotation-invariant; we show that this fails, predictably and by an exact amount, when capacity is limited. A symmetry groups a model's directions into blocks of equally strong directions, a symmetric model must use each block whole, and when the budget cuts a block every optimal model breaks the symmetry. For a Gaussian problem with a $k$-unit bilinear predictor, the log-loss price of symmetry is exactly $(U_k-I_k)/(2\sigma^2)$, the signal of the best $k$ directions minus that of the best whole blocks. Because an attention head compares tokens through a similarity of rank at most its head dimension, whatever computation precedes it, the same price governs heads fitted to a symmetric similarity by squared error, and heads trained this way on three-dimensional directions break symmetry at exactly the predicted head dimensions, with the predicted rotation sensitivity (RMSE 0.004). Under softmax attention, flatter attention restores the predicted symmetric head where tested, while with peaked attention broken heads beat every whole-shell symmetric head at 16 of 20 head dimensions. We give a procedure that separates capacity-driven breaking from other causes.
\end{abstract}

\section{Introduction}\label{sec:intro}

Many prediction problems have symmetries: rotating a molecule does not change its energy, and relabelling an unordered set does not change the set. Trained models often violate them, giving different answers for equivalent inputs. The usual explanations point at something more effort would fix: asymmetric data~\citep{lawrence2026augment}, an architecture that cannot represent the symmetric function, or unfinished training.

We isolate a cause that none of these remedies remove: limited capacity. Even with unlimited symmetric data and exact optimisation, a model with finite capacity can do strictly better by breaking a symmetry that the task genuinely has. The best models then form a \emph{family}: every transformed copy of a best model is also best. The symmetry survives in the set of solutions but not in any single one, which physicists call spontaneous symmetry breaking.

\begin{figure}[t]
  \centering
  \includegraphics[width=\columnwidth]{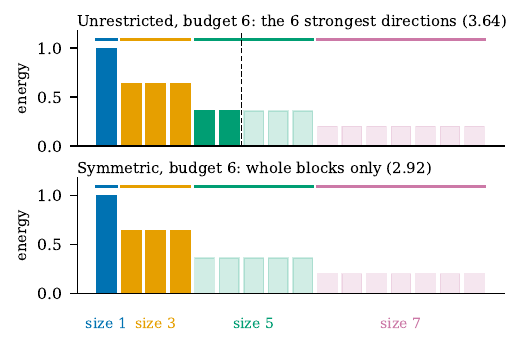}
  \caption{\textbf{Why limited capacity breaks symmetry.} Bars are directions, coloured by block (sizes 1, 3, 5, 7). With budget 6, the unrestricted model keeps the 6 strongest directions and cuts the third block; the symmetric model must keep whole blocks and leaves two units unused. The signal it gives up, $3.64-2.92$, divided by $2\sigma^2$, is the exact log-loss price of symmetry (\cref{thm:main}).}
  \label{fig:blocks}
\end{figure}

The mechanism is simple. A symmetry groups the directions a model can use into blocks that it mixes; the three coordinate axes of space, for example, form a block under rotations. A symmetric model cannot prefer one direction of a block over another, so it must keep or drop each block whole. In linear-algebra terms, the symmetry ties the target's eigenvalues within each block, the best $k$-unit model keeps the $k$ largest eigenvalues, and when $k$ falls inside a tied group every choice of directions from it is optimal but none is symmetric (\cref{fig:blocks}).

Attention heads are a direct instance. Whatever computation produces queries and keys, a head compares two tokens through $k$ numbers, so its similarity has rank at most its head dimension~\citep{bhojanapalli2020low}. A head that must learn a symmetric similarity faces exactly this budget problem.

\paragraph{Contributions.}
\begin{itemize}
  \item \textbf{An exact price} (\cref{sec:theorem}): for a Gaussian problem with a $k$-unit bilinear predictor, staying symmetric costs $(U_k-I_k)/(2\sigma^2)$; whenever this is positive, every optimal predictor is broken.
  \item \textbf{Attention heads} (\cref{sec:attention}): the same price governs heads fitted to a symmetric similarity by squared error on their logits, with head dimension as the budget.
  \item \textbf{Experiments} (\cref{sec:experiments}): heads trained by logit regression break symmetry at exactly the predicted head dimensions, with the predicted rotation sensitivity. Under softmax attention, flatter attention restores the predicted symmetric head where tested, and with peaked attention broken heads beat the best symmetric head built from whole shells at most head dimensions.
  \item \textbf{A procedure} (\cref{sec:method}) for deciding whether a symmetry violation in a trained model is the price of capacity.
\end{itemize}

\section{The idea: blocks and budgets}\label{sec:concepts}

A task is \emph{symmetric} under a group $G$ of input transformations when
\begin{equation}\label{eq:invariance}
  (gX,\,Y) \overset{d}{=} (X,\,Y)\qquad\text{for every } g\in G .
\end{equation}
If a model class contains every transformed copy of its members, the loss cannot tell a model from its copies, so the set of best models is symmetric (\cref{prop:orbit}); a unique best model is symmetric, but non-unique best models can each be broken. If the class is also closed under averaging over the symmetry, a symmetric best model exists, because averaging never increases log loss (\cref{thm:closure}). Breaking therefore needs a class in which the symmetric average costs more than the model it came from.

\paragraph{Blocks.} A symmetry mixes a model's directions in groups. Rotations mix the three linear functions $x_1,x_2,x_3$ of a direction in space, and functions of a direction split into blocks of sizes $1,3,5,7,\dots$, the spherical harmonics of each degree~\citep{atkinson2012spherical}. A symmetric model must give every direction of a block the same weight, because the symmetry mixes the block so thoroughly that only equal weights survive every transformation.\footnote{This is Schur's lemma: a linear map commuting with the symmetry acts on each block, an irreducible representation, as a multiple of the identity.} For example, the cosine $x\cdot y=x_1y_1+x_2y_2+x_3y_3$ of two unit vectors is rotation-invariant and needs three products. With two products the best one can do is $x_1y_1+x_2y_2$ or a rotated copy, which ignores an axis; a rotation-invariant model with two products can only output a constant.

\paragraph{Budgets and the price.} Give a model $k$ units, each carrying one direction. Let $U_k$ be the signal captured by the $k$ strongest individual directions and $I_k$ the signal captured by the best whole blocks of total size at most $k$, a knapsack value. The unrestricted model achieves $U_k$, the symmetric one $I_k$, and $U_k-I_k$ is the price of symmetry. It is zero when the $k$ strongest directions form whole blocks and positive in between. In linear-algebra terms this is a truncated eigendecomposition: the symmetry ties the target's eigenvalues within each block, and the best $k$-unit model keeps the $k$ largest, as a truncated SVD does~\citep{eckart1936approximation}. If $k$ falls inside a tied block, the model must choose some of its directions; every choice is equally good, none is symmetric, and rotating a choice gives another.

\paragraph{Rotation sensitivity}\label{sec:sensitivity} measures how broken a model is: the mean squared change in its output when a uniformly random symmetry transformation is applied to the input (to both items, for a model that compares two). It is zero for symmetric models.

\paragraph{Which budgets count.}\label{sec:budget} The budget must be one a symmetric model spends block by block: the model sees the relevant structure only through $k$ bilinear features, as when two items are compared through a $k$-dimensional inner product. Width spent on one input at a time does not count, since a single unit can compute an invariant of one input, such as a vector's length (\cref{rem:rank}). A comparison of two inputs, such as their angle, must pass through the shared interface.

\section{The exact price of staying symmetric}\label{sec:theorem}

\subsection{Construction}\label{sec:construction}

Let $\rho:G\to O(V)$ be an orthogonal representation of a finite or compact group $G$ on a $d$-dimensional real space $V$, acting on symmetric matrices by conjugation,
\begin{equation}\label{eq:conjugation}
  g\cdot X=\rho(g)\,X\,\rho(g)^\top .
\end{equation}
Let $X$ be a standard isotropic Gaussian in the space $\Sym(V)$ of symmetric matrices: in any Frobenius-orthonormal basis its coordinates are independent $N(0,1)$. For $M\in\Sym(V)$, consider the fixed-variance predictor
\begin{equation}
  q_M(\,\cdot\mid X)=N\!\left(\inner{M}{X}_F,\ \sigma^2\right)
\end{equation}
and the capacity-$k$ class
\begin{equation}\label{eq:class}
  \F_k=\{\,q_M:\ M\succeq 0,\ \rank(M)\le k\,\}.
\end{equation}
Equivalently $M=UU^\top$ with $U\in\R^{d\times k}$, and the mean is a sum of $k$ bilinear units,
\begin{equation}\label{eq:units}
  m_U(X)=\tr(U^\top X U)=\sum_{j=1}^{k} u_j^\top X u_j .
\end{equation}
If $X=(xy^\top+yx^\top)/2$ is built from a query $x$ and a key $y$, then $m_U(X)=(U^\top x)\cdot(U^\top y)$, the logit of a linear attention head with tied weights and head dimension $k$. Conjugation preserves positive semidefiniteness and rank, so $\F_k$ contains the transformed copies of its members, but it is \emph{not} closed under averaging, which can raise the rank.

\paragraph{Blocks.} Assume $V=\bigoplus_{a=1}^{m}V_a$, $d_a=\dim V_a$, splits into pairwise inequivalent, absolutely irreducible real representations.\footnote{Absolutely irreducible: irreducible over $\mathbb{C}$, so the only maps commuting with the symmetry on a block are multiples of the identity. Pairwise inequivalent: no two blocks transform alike.} By Schur's lemma the symmetric matrices commuting with every $\rho(g)$ are exactly the block scalars
\begin{equation}\label{eq:commutant}
  \textstyle M=\bigoplus_{a=1}^{m}c_aI_{V_a},\qquad c_a\in\R .
\end{equation}

\paragraph{Task.} Let
\begin{equation}\label{eq:target}
  \textstyle M_\star=\bigoplus_{a=1}^{m}\lambda_aI_{V_a},\qquad \lambda_a>0,
\end{equation}
and generate labels by
\begin{equation}\label{eq:population}
  Y\mid X\sim N\!\left(\inner{M_\star}{X}_F,\ \sigma^2\right).
\end{equation}
The target weighs all directions of a block equally, so the joint law of $(X,Y)$ is $G$-invariant. Collect the signal energies in the multiset $\Lambda=\{\lambda_a^2$ repeated $d_a$ times$\}$; let $U_k$ be the sum of its $k$ largest entries and
\begin{equation}\label{eq:knapsack}
  I_k=\max_{J:\ \sum_{a\in J}d_a\le k}\ \sum_{a\in J}d_a\lambda_a^2 .
\end{equation}

\subsection{The theorem}

For a class $\F$ with symmetric members $\F^G$ and population log loss $R_P$, the price of staying symmetric is
\begin{equation}\label{eq:gap}
  \delta_G(\F;P)=\inf_{f\in\F^G}R_P(f)-\inf_{f\in\F}R_P(f).
\end{equation}

\begin{theorem}[Exact block-completion gap]\label{thm:main}
For the population \eqref{eq:population} and class \eqref{eq:class}, with $0\le k\le d$,
\begin{equation}\label{eq:main}
  \boxed{\ \delta_G(\F_k;P)=\frac{U_k-I_k}{2\sigma^2}.\ }
\end{equation}
The unrestricted rank budget captures individual spectral directions, while an invariant coefficient can activate a block $V_a$ only by paying its full rank $d_a$. Hence strict population symmetry breaking occurs exactly when $U_k>I_k$.
\end{theorem}

\begin{proof}
Isotropy makes $g\cdot X\overset{d}{=}X$. Since $M_\star$ commutes with $\rho(G)$, $\inner{M_\star}{g\cdot X}_F=\inner{M_\star}{X}_F$, which proves exact joint invariance. For any $M\succeq0$,
\begin{equation}\label{eq:frob}
\begin{aligned}
  R_P(q_M)-R_P(q_{M_\star})&=\frac{1}{2\sigma^2}\E\!\left[\inner{M-M_\star}{X}_F^2\right]\\
  &=\frac{\|M-M_\star\|_F^2}{2\sigma^2}.
\end{aligned}
\end{equation}
Eckart--Young truncation of the positive spectrum~\citep{eckart1936approximation} therefore gives unrestricted captured energy $U_k$. Because the Gaussian design has full support, $q_M$ is invariant only if its two linear mean functionals agree everywhere, equivalently if $M$ commutes with $\rho(G)$. By \eqref{eq:commutant}, every invariant feasible coefficient is $\bigoplus_ac_aI_{V_a}$ with $c_a\ge0$. A nonzero $c_a$ consumes rank $d_a$, and conditional on activating the block its squared-error optimum is $c_a=\lambda_a$. The best invariant captured energy is therefore exactly the knapsack value $I_k$. Subtracting the two minimum risks in \eqref{eq:frob} proves \eqref{eq:main}.
\end{proof}

In words: with Gaussian noise, log loss is squared error, and the best rank-$k$ coefficient truncates the eigendecomposition of $M_\star$, capturing $U_k$. The symmetry ties eigenvalues within blocks, and a symmetric coefficient must use blocks whole, capturing at most $I_k$. When $k$ falls inside a tied block, every optimal coefficient keeps part of it and is broken, and its transformed copies are equally optimal.

\subsection{Simple cases}

\begin{corollary}[One block: every optimum picks a subspace]\label{cor:onefamily}
Let $G=O(d)$ act in its standard representation and $M_\star=\lambda I_d$ with $0<k<d$. Then $\delta_G(\F_k;P)=k\lambda^2/(2\sigma^2)$. The only invariant feasible coefficient is $M=0$, while every unrestricted minimiser is $\lambda P_W$, with $P_W$ the orthogonal projection onto a $k$-dimensional subspace $W$. The minimisers form a single $O(d)$-orbit of dimension $k(d-k)$, and none is invariant.
\end{corollary}

\begin{corollary}[Sawtooth capacity law]\label{cor:sawtooth}
If every block has dimension $r$ and $\lambda_1\ge\cdots\ge\lambda_m>0$, then for $k=qr+s$ with $0\le s<r$ and $q<m$, $\delta_G(\F_k;P)=s\lambda_{q+1}^2/(2\sigma^2)$: zero at completed blocks, rising linearly within the next block.
\end{corollary}

\begin{proposition}[Exact orbit defect]\label{prop:defect}
In \cref{cor:onefamily}, let $M=\lambda P_W$ and draw $g$ uniformly at random from $O(d)$ (Haar measure), independently of $X$. Then the rotation sensitivity is
\begin{equation}
  \E_{g,X}\!\left[\big(\inner{M}{X}_F-\inner{M}{g\cdot X}_F\big)^2\right]=2\lambda^2k\Big(1-\frac{k}{d}\Big).
\end{equation}
\end{proposition}

\begin{remark}[What the theorem calls rank]\label{rem:rank}
The constraint is the rank of the reshaped bilinear coefficient $M=UU^\top$, not the rank of an ordinary scalar-output linear map, which for $\inner{M}{X}_F$ is always one. A sum of $k$ bilinear units selects only $k$ eigendirections, and averaging over the symmetry can raise this structured rank. In ordinary deep-linear MSE models, invariant low-rank truncation can remain invariant~\citep{duan2026invariance}. Nor does generic hidden width alone force the gap: an architecture with a trace-pooling path computes $\tr(X)$ through one scalar.
\end{remark}

\paragraph{Beyond the Gaussian construction.}\label{sec:repair} \Cref{app:theory} proves that every optimum is again broken for a Bernoulli label (\cref{thm:bernoulli}); that the formula extends exactly to repeated blocks (\cref{thm:isotypic}); that the broken phase survives small symmetric distortions of the input (\cref{prop:aniso}); and that restoring symmetry by averaging costs capacity: the average of the broken predictor of \cref{cor:onefamily} has full rank, and distilling it into a student of rank $\ell<d$ breaks the symmetry again, at an exact price (\cref{thm:distill}).

\section{Attention heads are rank bottlenecks}\label{sec:attention}

\paragraph{A head is a rank-$k$ similarity.} Let tokens carry features $x\in\X$ drawn from $\mu$, raw or produced by any per-token network. A head computes $\phi_Q(x),\phi_K(y)\in\R^k$ and scores
\begin{equation}\label{eq:logit}
  s(x,y)=\phi_Q(x)^\top\phi_K(y)=\textstyle\sum_{t=1}^{k}\phi_{Q,t}(x)\,\phi_{K,t}(y).
\end{equation}
For $n$ queries and keys stacked as the rows of $Q$ and $K$, the logit matrix $QK^\top$ has rank at most $k$, however $Q$ and $K$ were computed.

\begin{lemma}[A head is a rank-$k$ similarity]\label{lem:rank}
For maps $\phi_Q,\phi_K:\X\to\R^k$ with coordinates in $L^2(\mu)$, the integral operator of $s$ has rank at most $k$, and so does the matrix of similarities on any finite set of queries and keys. Conversely, every square-integrable similarity of rank at most $k$ arises this way.
\end{lemma}

Per-token computation can reshape features arbitrarily, but a pairwise comparison passes through $k$ numbers and cannot be precomputed one token at a time (\cref{sec:budget}).\label{sec:attention-rank}

\paragraph{Symmetric similarities come in blocks.}\label{sec:attention-blocks} Let $G$ act on token features, jointly on query and key, and let the target similarity satisfy $\kappa(gx,gy)=\kappa(x,y)$. The blocks can be seen directly: the eigenvalues of the matrix $[\kappa(x_i,x_j)]/n$ on $n$ sampled tokens fall into tied groups whose sizes are the block sizes. For directions in $\R^3$ and the similarity of our experiments, 3,000 tokens give groups of sizes 1, 3, 5, 7, 9, 11 around the predicted values, up to sampling error. These blocks are the spherical harmonics of degree $l$, which we call \emph{shells}.

\paragraph{The exact head-dimension gap.}\label{sec:attention-gap} Let a compact group $G$ act continuously on $\X$, let $\mu$ be $G$-invariant, and draw $(x,y)\sim\mu\otimes\mu$. Let $\kappa$ be square-integrable, symmetric and $G$-invariant, with integral operator $T_\kappa=\sum_a\lambda_aP_a$, where $\lambda_a\neq0$ and $P_a$ projects onto $V_a$; the $V_a$ are mutually orthogonal, pairwise inequivalent, absolutely irreducible representations with $d_a=\dim V_a$. Observe $Y=\kappa(x,y)+\varepsilon$ with $\varepsilon\sim N(0,\sigma^2)$, and let $\Hd_k$ be the heads \eqref{eq:logit} with $\phi_Q,\phi_K$ in $L^2(\mu)$, each predicting $N(s(x,y),\sigma^2)$. Call a head symmetric when $s(gx,gy)=s(x,y)$ almost everywhere for every $g$, and define $U_k$ and $I_k$ as in \cref{sec:construction}.

\begin{theorem}[Exact head-dimension gap]\label{thm:attention}
Under these assumptions, for every head dimension $k$,
\begin{equation}\label{eq:attention}
  \delta_G(\Hd_k;P)=\frac{U_k-I_k}{2\sigma^2}.
\end{equation}
When $U_k>I_k$, every optimal head breaks the symmetry, and every transformed copy $s(g\,\cdot,g\,\cdot)$ of an optimal head is optimal.
\end{theorem}

\begin{proof}[Proof sketch]
The excess log loss is $\E[(s-\kappa)^2]/(2\sigma^2)$, the squared Hilbert--Schmidt (Frobenius-like) distance between $T_s$ and $T_\kappa$. By \cref{lem:rank} heads are exactly the rank-$\le k$ operators, so the best head keeps the $k$ largest-magnitude eigendirections~\citep{eckart1936approximation,mirsky1960symmetric} and captures $U_k$. A symmetric head commutes with the symmetry, so it must treat every direction of a block alike and spend the block's full dimension to use it, which gives $I_k$. \Cref{app:attention} gives the full proof.
\end{proof}

In words: a head that must learn a symmetric similarity faces the budget problem of \cref{sec:theorem}, with head dimension as the budget. The similarity's eigenvalues are tied within blocks, and the best head keeps the $k$ strongest eigendirections; when $k$ cuts a block, it keeps part of it and breaks the symmetry. No per-token computation changes this, because the bottleneck is the $k$-dimensional interface between query and key.

\begin{corollary}[Rotation-invariant similarity on the sphere]\label{cor:sphere}
Let $\X=S^2$ with the uniform distribution, $G=SO(3)$, and
\begin{equation}\label{eq:sphere-target}
\begin{gathered}
  \kappa(x,y)=\textstyle\sum_{l=0}^{L}\lambda_l\,(2l+1)\,P_l(x\cdot y),\\
  \lambda_0>\lambda_1>\cdots>\lambda_L>0,
\end{gathered}
\end{equation}
with $P_l$ the Legendre polynomial of degree $l$, a polynomial in the cosine. For $1\le k\le(L+1)^2$, a symmetric optimal head exists exactly when $k\in\{1,4,9,\dots\}$. For $k=l^2+r$ with $0<r<2l+1$, every optimal head is broken and, for independent $x$, $y$ and uniformly random rotation $g$,
\begin{equation}\label{eq:sphere-dispersion}
  \E\!\left[\big(s(x,y)-s(gx,gy)\big)^2\right]=2\lambda_l^2\,r\Big(1-\frac{r}{2l+1}\Big).
\end{equation}
\end{corollary}

\begin{figure*}[t]
  \centering
  \includegraphics[width=0.92\textwidth]{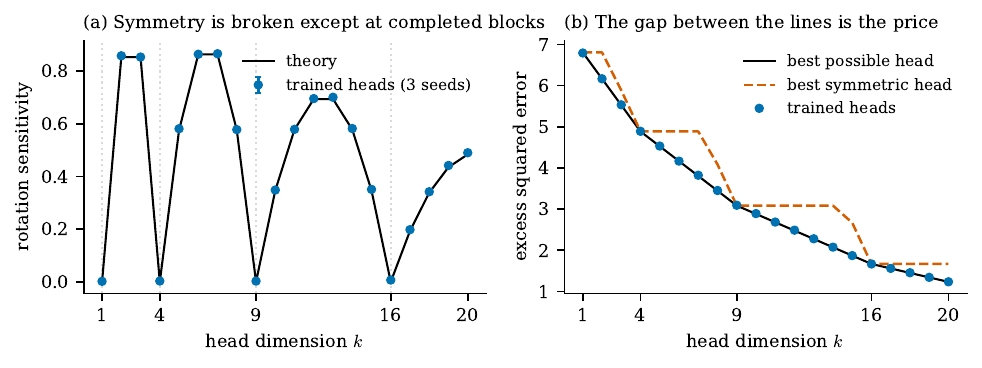}
  \caption{\textbf{Logit regression reproduces \cref{cor:sphere}.} (a) Rotation sensitivity of trained heads (mean $\pm$ s.d., three seeds) against the prediction \eqref{eq:sphere-dispersion}; heads are symmetric only where $k$ completes a shell (dotted). (b) Excess squared error of trained heads, of the best head ($S-U_k$, with $S=\sum_ad_a\lambda_a^2$) and of the best symmetric head ($S-I_k$). The gap between the lines is the price trained heads avoid by breaking.}
  \label{fig:logit}
\end{figure*}

\paragraph{What changes under softmax.}\label{sec:softmax-theory} A real head applies a softmax over keys and is trained by cross-entropy. First, functions of the query alone do not affect attention, so we call a softmax head symmetric when its logits centred over keys are invariant. Centring removes the constant shell and keeps rank $\le k$, so symmetric heads still use whole blocks, but the constant one is free: on the sphere, the strongest shells $l=1,\dots,j$ fill head dimensions $3,8,15,\dots$. Second, when target and head logits are small (flat attention), the KL between attention distributions is approximately half the variance over keys of the logit difference. In this limit \cref{thm:attention} applies to centred logits, and the best heads on the sphere are symmetric exactly at $k=3,8,15,\dots$. Third, under peaked attention the loss is not quadratic, the value of a direction depends on which others the head has, and whether breaking beats every whole-block head is empirical. Our results concern the head that performs a comparison; in deep models earlier layers can move information between tokens and route around it.

\begin{figure*}[t]
  \centering
  \includegraphics[width=0.92\textwidth]{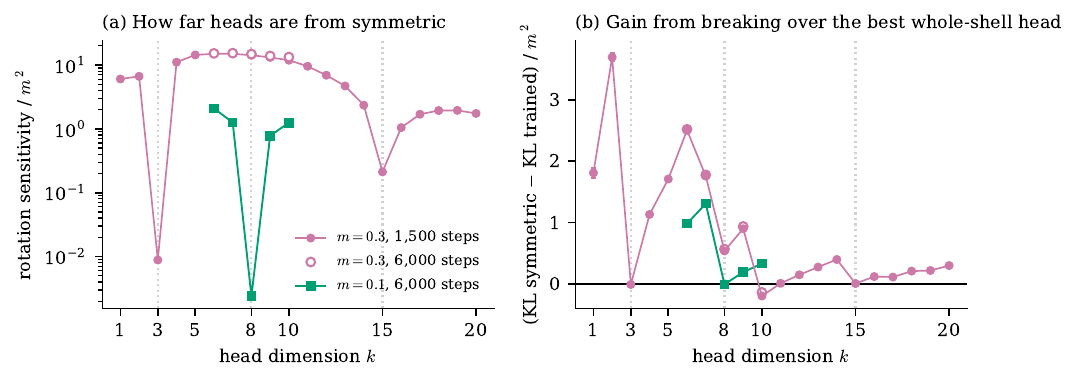}
  \caption{\textbf{Softmax attention.} Dotted lines mark the predicted symmetric heads ($k=3,8,15$); both panels divide by $m^2$. (a) Rotation sensitivity (means over three seeds). At $m=0.1$ the head at $k=8$ is symmetric, over 300 times less sensitive than its neighbours; at $m=0.3$ heads are symmetric at $k=3$, nearly so at $15$, and broken at $8$. (b) KL of the best whole-shell head ($l\le5$) minus KL of trained heads (s.d.\ bars on the 1,500-step series); positive means breaking wins. The negative values at $k=10$, $m=0.3$ are where training ends at a broken head that a symmetric head outperforms (\cref{sec:spread}).}
  \label{fig:softmax}
\end{figure*}

\section{Diagnosing capacity-driven breaking}\label{sec:method}

The theory turns ``why is my model not symmetric?'' into a test that needs only measurable quantities, not the Gaussian assumptions.

\begin{methodbox}{Procedure: is a symmetry violation the price of capacity?}
\begin{enumerate}[leftmargin=1.2em]
  \item \textbf{Symmetry and data.} State how the symmetry acts (on a query and all its keys together) and check that the data are invariant; if not, asymmetric data compete as an explanation.
  \item \textbf{Budget.} Find the narrowest pairwise inner product: head dimension, dual-encoder embedding size, or bilinear rank. Its inputs must carry the symmetry.
  \item \textbf{Blocks.} Look up block sizes (\cref{tab:blocks-main}) or find tied groups of eigenvalues of the target or a large model's similarity on sampled inputs.
  \item \textbf{Rotation sensitivity.} Apply a random transformation to each input and record the mean squared output change, centring logits over keys for softmax. Divided by twice the mean squared (centred) logit, it lies in $[0,1]$.
  \item \textbf{Sweep the budget.} Capacity-driven breaking drops by orders of magnitude where whole blocks fill the budget; non-quadratic losses can blur this.
  \item \textbf{Best symmetric model.} Fit the best model built from whole blocks at each budget, e.g.\ logits $\sum_{l\in J}c_l(2l+1)P_l(x\cdot y)$ on the sphere. Breaking is the price of capacity only if the trained model's loss is lower.
  \item \textbf{Locate the cut} (optional). Measure how many dimensions of each block the features use; partly used blocks are cut.
\end{enumerate}
\end{methodbox}

\Cref{app:practice} tabulates how to read each outcome; \cref{tab:blocks-main} gives block sizes for common symmetries. If breaking is the price of capacity, there are four options:
\begin{itemize}
  \item \textbf{Complete the blocks}: raise the budget until $U_k=I_k$. This removes the price under squared error, but not under peaked softmax (\cref{sec:softmax-results}).
  \item \textbf{Make the symmetric answer cheap} with invariant inputs, such as distances or angles, or equivariant layers~\citep{cohen2016group,thomas2018tensor,fuchs2020se3}.
  \item \textbf{Average and distil with room}: averaging restores symmetry without raising the loss, but the average uses every block the broken model touched whole, and a smaller student can break again (\cref{thm:distill}).
  \item \textbf{Leave it}: the broken model has the lower loss; repair pays only when consistency on equivalent inputs matters more.
\end{itemize}

\begin{table}[t]
\centering\footnotesize
\caption{Blocks of common symmetries and the budgets at which whole blocks fit, assuming strengths decrease in the order listed; under softmax the constant block is free (bracketed).}\label{tab:blocks-main}
\setlength{\tabcolsep}{3pt}
\begin{tabular}{@{}lll@{}}
\toprule
Items, symmetry & Block sizes & Whole-block budgets\\
\midrule
Angles, $O(2)$ & $1,2,2,\dots$ & $1,3,5,\dots$ \ ($2,4,\dots$)\\
Directions in $\R^3$, $SO(3)$ & $1,3,5,\dots$ & $1,4,9,\dots$ \ ($3,8,\dots$)\\
Directions in $\R^d$, $SO(d)$ & $1,\,d,\,\tfrac{d(d+1)}{2}{-}1,\dots$ & $1,\,d{+}1,\dots$ \ ($d,\dots$)\\
$n$ symbols, $S_n$ & $1,\,n{-}1$ & $1,\,n$ \ ($n{-}1$)\\
\bottomrule
\end{tabular}
\end{table}

\section{Experiments}\label{sec:experiments}

\paragraph{Setup.} Tokens are uniform unit vectors in $\R^3$. The target is \eqref{eq:sphere-target} with $L=5$ and $\lambda=(1,0.8,0.6,0.45,0.33,0.25)$, so blocks have sizes $1,3,\dots,11$. Queries and keys come from separate SiLU MLPs, $3\to128\to128\to128\to k$, so per-token computation is expressive and only $k$ is limited. \emph{Logit regression} fits $s$ to $\kappa$ by squared error (\cref{thm:attention}). \emph{Softmax attention} minimises the KL to $\mathrm{softmax}(m\kappa)$ over 64 keys; $m=0.3$ is peaked (perplexity about 24 of 64 keys) and $m=0.1$ flatter (about 60). Before training we recorded the predicted symmetric dimensions, $k=1,4,9,16$ for logit regression and $k=3,8,15$ for softmax, and the predicted sensitivity \eqref{eq:sphere-dispersion}. Models are trained on fresh samples with Adam, three seeds, 1,500 steps for $k=1,\dots,20$ and 6,000 steps in a follow-up at $k=6,\dots,10$; for softmax at $m=0.3$ and $k=6,\dots,10$, quadrupling the steps changed the mean KL by at most 0.005. Every run is reported at its final step (\cref{app:experiments}).

\paragraph{Logit regression: the theorem holds.}\label{sec:logit-results} Trained heads are symmetric at $k=1,4,9,16$ (sensitivity 0.002--0.007) and broken elsewhere (0.20--0.87). The sensitivity matches \eqref{eq:sphere-dispersion} with RMSE 0.0041 across $k=1,\dots,20$, and the loss is within about 0.02 of the optimum $S-U_k$, well below the best symmetric head wherever $k$ cuts a shell (\cref{fig:logit}). Which orientation a broken head picks is left to training; with three seeds we cannot test whether it is uniform, and Adam, which rescales weights separately, is not rotation-equivariant (\cref{app:experiments}).

\paragraph{Softmax attention.}\label{sec:softmax-results} With flatter attention the prediction holds over the dimensions tested: the head at $k=8$ is symmetric (scaled sensitivity 0.0024 against 0.78--2.1 for $k=6,7,9,10$) and matches the best whole-shell head, while elsewhere breaking wins (\cref{fig:softmax}). With peaked attention, heads are symmetric at $k=3$ and nearly so at $k=15$, but at $k=8$ the prediction fails: heads break (scaled sensitivity 14.8 after 6,000 steps) and beat the best whole-shell head (KL 0.208 against 0.259). Across the sweep, broken heads beat every whole-shell head at 16 of 20 dimensions, by 0.010 to 0.33 in KL; they tie at $k=11$, and at $k=10$ they lose (0.150 against 0.138), so there the breaking is not the price of capacity (\cref{tab:softmax}).

\begin{table}[t]
\centering\footnotesize
\caption{Softmax attention: trained heads against the best head built from whole shells $l\le5$ (KL to the target attention; lower is better; means over three seeds, 6,000 steps for $k=8,10$ and 1,500 otherwise). Sensitivity is divided by $m^2$.}\label{tab:softmax}
\setlength{\tabcolsep}{3pt}
\begin{tabular}{@{}ccllll@{}}
\toprule
$m$ & $k$ & Whole shells & Trained & Sens. & Reading\\
\midrule
0.3 & 3  & 0.566 (1)     & 0.567 & 0.009 & symmetric\\
0.3 & 8  & 0.259 (1,2)   & 0.208 & 14.8  & breaking wins\\
0.3 & 10 & 0.138 (1,3)   & 0.150 & 13.3  & symmetric better\\
0.3 & 11 & 0.138 (1,3)   & 0.137 & 9.6   & tie\\
0.3 & 15 & 0.091 (1,2,3) & 0.091 & 0.21  & nearly symmetric\\
0.1 & 8  & 0.0228 (1,2)  & 0.0228 & 0.0024 & symmetric\\
0.1 & 10 & 0.0222 (1,3)  & 0.0189 & 1.24  & breaking wins\\
\bottomrule
\end{tabular}
\end{table}

\paragraph{Why peaked attention breaks more.}\label{sec:spread} Replacing the MLPs by exact spherical-harmonic features up to degree 5, so that every shell is equally easy to represent, leaves the breaking intact: trained heads spread over several shells, each partly used (\cref{tab:spread}). The objective, not the network, is responsible. Under squared error the values of directions add, so optima fill shells in order; under peaked softmax they do not, and even the ranking of single shells reverses (KL 0.407, 0.526, 0.566 for shells 3, 2, 1 alone). At $k=8$ and $15$ the spread heads beat every whole-shell head (0.208--0.212 against 0.259; 0.078--0.081 against 0.091), so breaking is the price of capacity there. At $k=10$ they do not (0.150--0.153 against 0.138); the symmetric head on shells 1 and 3 is stable when started nearby, but training from random weights does not find it. The MLP heads look symmetric at $k=15$ only because they barely use shells 4 and 5 (under 0.5\% of feature variance), consistent with spectral bias~\citep{rahaman2019spectral}.

\paragraph{Summary.} Under squared error, limited head dimension breaks symmetry exactly as \cref{thm:attention} predicts: at the predicted dimensions, with the predicted sensitivity and gain. Under softmax, flatter attention restores the predicted symmetric head where tested, and peaked attention spreads capacity across blocks, so that breaking beats every whole-shell head at most dimensions. Where it does not ($k=10$), step 6 of the procedure identifies the breaking as a training outcome rather than a price of capacity.

\section{Applicability, related work and limits}\label{sec:applicability}

\paragraph{Where it applies.} The rank bound needs only that two items meet in a $k$-dimensional inner product: attention heads, dual encoders~\citep{karpukhin2020dense}, contrastive embeddings~\citep{radford2021learning} and low-rank bilinear layers. In dual encoders nothing routes around the inner product; in deep transformers, earlier layers can move information between tokens, so the result concerns the head that performs a comparison. The exact price additionally needs squared error and independent symmetric inputs; elsewhere the procedure decides. The symmetry must act on the pair: if the score is invariant for each item separately, each side computes its invariants alone and pays nothing.

\paragraph{Does it matter at scale?} The price is at most $r\lambda^2/(2\sigma^2)$ when the budget affords $r$ directions of a block of energy $\lambda^2$ each, so it matters when strong blocks are cut. For directions in $\R^3$, heads of dimension 64 or 128 fit all shells up to degree 7 or 10, and a smooth similarity carries little signal beyond. But blocks grow fast with input dimension: for directions in $\R^{16}$ they have sizes $1,16,135,800$, so a symmetric 64-dimensional head can represent the cosine but no part of the quadratic block. A head matching token identity over $n$ equally likely tokens faces a single block of size $n-1$; any useful head of dimension below $n-1$ must favour some tokens over others (\cref{app:practice}). Natural candidates are attention over atoms, point clouds and physical states, where heads compare orientations.

\paragraph{Related work.} \emph{The cost of enforcing symmetry.} \citet{chen2026drawback} show that enforcing equivariance can reduce the expressive power of fixed-width two-layer ReLU networks and quantify the width needed to compensate; we do not claim the general statement that symmetry constraints can cost expressivity, but give an exact price for a structured-rank family, its broken family of optima, its extension to attention and an exact distillation gap. \citet{maass2024symmetries} separate weakly and strongly invariant solutions in overparameterised networks. \emph{Averaging.} Orbit averaging, augmentation and Reynolds operators give the positive boundary~\citep{chen2020group,elesedy2021provably,sannai2024invariant,orbanz2024global}: when averaging stays inside the class, a symmetric optimum exists (\cref{thm:closure}); our theorems quantify what happens when it does not. \emph{Low-rank models.} \citet{kohn2025geometry} solve block-rank allocations for cyclic permutation-equivariant linear models with blockwise Eckart--Young, and \citet{duan2026invariance} show that ordinary rank-bounded deep-linear models can keep invariant optima; \cref{rem:rank} explains why the representation on which rank is imposed decides the outcome. \emph{Attention.} \citet{bhojanapalli2020low} identify the low-rank bottleneck of attention as a limit on expressivity; we show that it forces symmetry breaking for symmetric similarities and price it exactly. \emph{Landscapes and distillation.} Symmetry-related minima arise in matrix factorisation~\citep{li2016symmetry} and tensor decomposition~\citep{arjevani2021symmetry}; here the orbit consists of global optima of a misspecified family, with a quantified risk penalty. Distillation compresses teachers into students~\citep{hinton2015distilling}, which often fail to inherit their teachers' invariances~\citep{odagiu2024knowledge,nix2022functional}; \cref{thm:distill} gives an exact population version. \emph{Other mechanisms.} Asymmetric data~\citep{lawrence2026augment}, fragile learned invariance~\citep{moskalev2023genuine} and emergent frequency bottlenecks~\citep{wen2024frequencies} are distinct causes; \citet{goel2026anysubgroup} break symmetry deliberately to select subgroups.

\paragraph{Limitations.} \Cref{thm:main} assumes an isotropic Gaussian design, a correctly specified fixed-variance Gaussian conditional and positive semidefinite coefficients; \cref{thm:attention} assumes a compact group, squared error on the logits, queries and keys drawn independently from a symmetric distribution, and a single head. Both need pairwise inequivalent, absolutely irreducible blocks; complex-type blocks, such as those of planar rotations without reflections, are not treated. For softmax attention our evidence is synthetic: one family of similarities on $S^2$, two multipliers, small networks and three seeds.

\section*{Impact Statement}

This paper explains when limited capacity forces models to treat equivalent inputs differently, and gives a procedure for diagnosing it. Its main use is in building more consistent models, for example in scientific applications with physical symmetries. We do not foresee specific negative societal consequences beyond those of machine learning research in general.

\bibliography{refs}
\bibliographystyle{plainnat}

\newpage
\appendix
\onecolumn
\input{appA_theory}
\input{appB_attention}
\input{appC_experiments}
\input{appD_practice}

\end{document}

%% file: appA_theory.tex
\section{Further theory and deferred proofs}\label{app:theory}

\subsection{Symmetric objectives and families of optima}

Let $\mu_G$ be the uniform probability measure when $G$ is finite and normalised Haar measure when $G$ is compact. For a conditional distribution $f$, define
\begin{equation}
  (T_gf)(y\mid x)=f(y\mid g^{-1}x),\qquad (\Pi_Gf)(y\mid x)=\int_G (T_gf)(y\mid x)\,d\mu_G(g).
\end{equation}
The class $\F$ is \emph{group-stable} if $T_gf\in\F$ for every $f\in\F$ and $g\in G$, and \emph{orbit-closed} if $\Pi_Gf\in\F$ for every $f\in\F$. Write $\F^G=\{f\in\F:\ T_gf=f\ P_X\text{-a.s. for every }g\in G\}$.

\begin{proposition}[Symmetric risk, unique optima, and degeneracy]\label{prop:orbit}
Assume $(gX,Y)\overset{d}{=}(X,Y)$, $\F$ is group-stable, and population log loss is finite at the candidates under consideration. Then
\begin{equation}\label{eq:symrisk}
  R_P(T_gf)=R_P(f)\qquad\text{for all } f\in\F,\ g\in G.
\end{equation}
Consequently the population minimiser set is $G$-stable. If the minimising predictor is unique as a conditional distribution, it is invariant. If both infima in \eqref{eq:gap} are attained and $\delta_G>0$, every minimiser has a proper stabiliser and belongs to a nontrivial equal-risk orbit
\begin{equation}
  G/\Stab(f),\qquad \Stab(f)=\{g:\ T_gf=f\}.
\end{equation}
If, additionally, a compact Lie group acts smoothly on a predictor manifold and the restricted risk is twice differentiable, the tangent space of the minimiser orbit lies in the kernel of the intrinsic Hessian on that predictor manifold.
\end{proposition}

\begin{proof}
The change of variables $X\mapsto g^{-1}X$ and joint invariance give \eqref{eq:symrisk}. Thus every transform of a minimiser is a minimiser. Uniqueness implies $T_gf=f$ for every $g$. When $\delta_G>0$, no minimiser is invariant, so its stabiliser is proper and its orbit is nontrivial. In the smooth case every point on the orbit is critical. Differentiating the zero intrinsic gradient along any one-parameter orbit shows that the intrinsic Hessian annihilates every orbit-tangent vector.
\end{proof}

\begin{theorem}[Orbit closure is the no-breaking boundary]\label{thm:closure}
Under the assumptions of \cref{prop:orbit},
\begin{equation}
  R_P(\Pi_Gf)\le R_P(f)
\end{equation}
for every full-support predictor $f$. If $\F$ is orbit-closed, then
\begin{equation}
  \inf_{f\in\F^G}R_P(f)=\inf_{f\in\F}R_P(f),\qquad \delta_G(\F;P)=0.
\end{equation}
\end{theorem}

\begin{proof}
Jensen's inequality for $-\log$ gives
\begin{equation}
  -\log\int_G f(Y\mid g^{-1}X)\,d\mu_G(g)\le\int_G -\log f(Y\mid g^{-1}X)\,d\mu_G(g).
\end{equation}
Taking expectations and using joint invariance makes the right side equal to $R_P(f)$. The left side is the risk of the invariant orbit mixture. If that mixture remains in $\F$, every unrestricted candidate has an invariant in-class competitor with no larger risk.
\end{proof}

\subsection{Repeated blocks}

\begin{theorem}[Repeated-isotypic extension]\label{thm:isotypic}
Let
\begin{equation}
  V=\bigoplus_{a=1}^{m}\left(W_a\otimes\R^{m_a}\right),\qquad \rho(g)=\bigoplus_{a=1}^{m}\left(\rho_a(g)\otimes I_{m_a}\right),
\end{equation}
where the $W_a$ are pairwise inequivalent absolutely irreducible real representations and $d_a=\dim W_a$. Replace \eqref{eq:target} by the invariant positive-semidefinite target
\begin{equation}
  M_\star=\bigoplus_{a=1}^{m}\left(I_{W_a}\otimes A_a\right),\qquad A_a\succeq0,
\end{equation}
and write the eigenvalues of $A_a$ as $\lambda_{a,1}\ge\cdots\ge\lambda_{a,m_a}\ge0$. Let $\widetilde U_k$ be the sum of the $k$ largest values in the multiset containing each $\lambda_{a,j}^2$ exactly $d_a$ times, and let
\begin{equation}
  \widetilde I_k=\max_{\substack{0\le r_a\le m_a\\ \sum_ad_ar_a\le k}}\ \sum_{a=1}^{m}d_a\sum_{j=1}^{r_a}\lambda_{a,j}^2 .
\end{equation}
Under the same isotropic Gaussian design and class $\F_k$,
\begin{equation}
  \delta_G(\F_k;P)=\frac{\widetilde U_k-\widetilde I_k}{2\sigma^2}.
\end{equation}
\end{theorem}

Thus repeated copies turn whole-block allocation into a weighted allocation of multiplicity-space eigenmodes; each selected mode still costs the full irreducible dimension $d_a$.

\begin{proof}
Real absolute irreducibility and pairwise inequivalence give the symmetric commutant
\begin{equation}
  \Sym(V)^G=\bigoplus_{a=1}^{m}\left(I_{W_a}\otimes\Sym(m_a)\right).
\end{equation}
The eigenvalues of $M_\star$ are those of $A_a$, each repeated $d_a$ times, so unrestricted rank-$k$ truncation captures $\widetilde U_k$. Every invariant feasible coefficient has the form $M=\bigoplus_a(I_{W_a}\otimes B_a)$ with $B_a\succeq0$, and
\begin{equation}
  \rank(M)=\sum_ad_a\rank(B_a),\qquad \|M-M_\star\|_F^2=\sum_ad_a\|B_a-A_a\|_F^2 .
\end{equation}
At multiplicity-space rank $r_a$, positive spectral truncation of $A_a$ captures its largest $r_a$ squared eigenvalues. Optimising the integers $r_a$ yields $\widetilde I_k$, and \eqref{eq:frob} completes the proof.
\end{proof}

\subsection{Stability under symmetric anisotropy}

\begin{proposition}[Stability under invariant anisotropy]\label{prop:aniso}
Retain the target and class of \cref{thm:main}, but let $X$ be a centred Gaussian in $\Sym(V)$ with covariance operator $\mathcal{C}$. Assume $\mathcal{C}$ commutes with the conjugation action and, for some $0\le\varepsilon<1$,
\begin{equation}
  (1-\varepsilon)I\preceq\mathcal{C}\preceq(1+\varepsilon)I .
\end{equation}
Let $S=\|M_\star\|_F^2=\sum_ad_a\lambda_a^2$ and write $\delta_G^{(\mathcal{C})}$ for the resulting approximation gap. Then
\begin{equation}\label{eq:aniso}
  \max\left\{0,\ \frac{U_k-I_k-\varepsilon(2S-U_k-I_k)}{2\sigma^2}\right\}\le\delta_G^{(\mathcal{C})}\le\frac{U_k-I_k+\varepsilon(2S-U_k-I_k)}{2\sigma^2}.
\end{equation}
In particular, the broken phase persists whenever $U_k-I_k>\varepsilon(2S-U_k-I_k)$.
\end{proposition}

\begin{proof}
Commutation of $\mathcal{C}$ with the orthogonal group action makes the centred Gaussian design invariant. For $D=M-M_\star$, excess log loss is $\inner{D}{\mathcal{C}D}_F/(2\sigma^2)$, which lies between $(1-\varepsilon)\|D\|_F^2/(2\sigma^2)$ and $(1+\varepsilon)\|D\|_F^2/(2\sigma^2)$. Under the Frobenius metric, the minimum invariant and unrestricted squared errors are $a=S-I_k$ and $b=S-U_k$, respectively. Hence the anisotropic difference of infima lies between $\{(1-\varepsilon)a-(1+\varepsilon)b\}/(2\sigma^2)$ and $\{(1+\varepsilon)a-(1-\varepsilon)b\}/(2\sigma^2)$. Using $a-b=U_k-I_k$, $a+b=2S-U_k-I_k$, and nonnegativity of the nested-class gap proves \eqref{eq:aniso}.
\end{proof}

\subsection{Proofs for the simple cases in \texorpdfstring{\cref{sec:theorem}}{Section 3}}

\begin{proof}[Proof of \cref{cor:onefamily}]
The invariant symmetric commutant of $O(d)$ consists of scalar matrices. No nonzero scalar matrix has rank below $d$. Unrestricted rank-$k$ approximation of $\lambda I_d$ retains $\lambda$ on an arbitrary $k$-plane. The value of the gap, transitivity of $O(d)$ on the set of $k$-planes, and the dimension $k(d-k)$ of that set follow.
\end{proof}

\begin{proof}[Proof of \cref{cor:sawtooth}]
With equal block sizes $r$ and ordered strengths, the $k=qr+s$ strongest directions are the first $q$ blocks and $s$ directions of block $q+1$, so $U_k=r\sum_{a\le q}\lambda_a^2+s\lambda_{q+1}^2$. At most $q$ whole blocks fit in the budget, and the strongest $q$ capture $I_k=r\sum_{a\le q}\lambda_a^2$. \Cref{thm:main} gives the result.
\end{proof}

\begin{proof}[Proof of \cref{prop:defect}]
Isotropy turns the conditional second moment into $\|M-\rho(g)^\top M\rho(g)\|_F^2$. Haar averaging sends $M$ to $\Pi_GM=(\tr M/d)I_d=\lambda(k/d)I_d$. Expanding the squared norm and averaging gives $2\|M\|_F^2-2\|\Pi_GM\|_F^2=2\lambda^2k-2\lambda^2k^2/d$.
\end{proof}

\subsection{A non-quadratic loss: Bernoulli prediction}

Fixed-variance Gaussian log loss gives the exact spectral formula because its mean regret is quadratic. The representation obstruction is not restricted to that geometry. Let $Z\in\R^{d\times d}$ have independent standard Gaussian entries, let $O(d)$ act by $Z\mapsto gZg^\top$, and define the invariant Bernoulli target
\begin{equation}\label{eq:bern-target}
  P(Y=1\mid Z)=\mathrm{sigm}(\lambda\tr Z),\qquad \mathrm{sigm}(t)=\frac{1}{1+e^{-t}},\qquad\lambda>0.
\end{equation}
For a fixed radius $M>0$, consider the $k$-channel bilinear class
\begin{equation}\label{eq:bern-class}
  \mathcal{B}_{k,M}=\left\{q_B(1\mid Z)=\mathrm{sigm}(\inner{B}{Z}_F):\ \rank(B)\le k,\ \|B\|_F\le M\right\},\qquad 1\le k<d .
\end{equation}
The factorisation $\inner{B}{Z}_F=\sum_{j=1}^{k}u_j^\top Zv_j$ makes $k$ a separable-channel budget.

\begin{theorem}[Bernoulli multiplet bottleneck]\label{thm:bernoulli}
Let $T=\tr Z\sim N(0,d)$ and define
\begin{equation}
  c_\lambda=\lambda\,\E\!\left[\mathrm{sigm}'(\lambda T)\right]>0 .
\end{equation}
Then the joint law induced by \eqref{eq:bern-target} is exactly $O(d)$-invariant and $\mathcal{B}_{k,M}$ is group-stable. Its only invariant predictor has $B=0$. Every global minimiser has the form
\begin{equation}\label{eq:bern-min}
  B^\star=t^\star P_W,\qquad W\in\mathrm{Gr}(k,d),\qquad t^\star>0,
\end{equation}
with a common scalar $t^\star$ determined by $(\lambda,d,k,M)$. In particular every optimum is non-invariant. If $M\ge4c_\lambda\sqrt{k}$, then
\begin{equation}\label{eq:bern-gap}
  \delta_{O(d)}(\mathcal{B}_{k,M};P)\ge2kc_\lambda^2>0 .
\end{equation}
\end{theorem}

The lower bound suffices to establish a strictly positive non-quadratic gap. It shows that the spectral Gaussian result is exposing a representation mechanism rather than relying on a quadratic scoring rule. The exact Gaussian theorem remains useful because it additionally yields the block-knapsack formula, sharp completion capacities and a closed-form distillation gap.

\begin{proof}
Gaussian isotropy under the Frobenius-orthogonal map $Z\mapsto gZg^\top$, together with trace invariance, proves exact joint symmetry. Conjugation preserves rank and Frobenius norm, so the class is group-stable. Since the logistic link is injective and the Gaussian design has full support, invariance of $q_B$ requires $gBg^\top=B$ for all $g\in O(d)$. The commutant consists of scalar matrices; $k<d$ therefore forces $B=0$.

Write the population Bernoulli log loss as
\begin{equation}\label{eq:bern-risk}
  R(B)=\E\!\left[\log\!\left(1+e^{\inner{B}{Z}_F}\right)-\mathrm{sigm}(\lambda T)\inner{B}{Z}_F\right].
\end{equation}
The pair $(\inner{B}{Z}_F,T)$ is jointly Gaussian with $\mathrm{Var}(\inner{B}{Z}_F)=\|B\|_F^2$ and $\mathrm{Cov}(\inner{B}{Z}_F,T)=\tr B$. Gaussian integration by parts therefore yields the exact reduction
\begin{equation}\label{eq:bern-reduce}
  R(B)=\Phi(\|B\|_F)-c_\lambda\tr B,\qquad \Phi(s)=\E_{G\sim N(0,1)}\log\!\left(1+e^{sG}\right).
\end{equation}
At fixed $b=\|B\|_F$ and rank at most $k$,
\begin{equation}\label{eq:trnuc}
  \tr B\le\|B\|_*\le\sqrt{k}\,b,
\end{equation}
with equality exactly for $B=(b/\sqrt{k})P_W$. Thus global minimisation reduces to the strictly convex one-dimensional problem
\begin{equation}
  \min_{0\le t\le M/\sqrt{k}}\left\{\Phi(t\sqrt{k})-kc_\lambda t\right\}.
\end{equation}
Its derivative at zero is $-kc_\lambda<0$, so $t^\star>0$, and the equality cases in \eqref{eq:trnuc} give \eqref{eq:bern-min}. Finally, along $B=tP_W$,
\begin{equation}
  R'(0)=-kc_\lambda,\qquad R''(t)=\E\!\left[\mathrm{sigm}'\!\left(\inner{tP_W}{Z}_F\right)\inner{P_W}{Z}_F^2\right]\le\frac{k}{4}.
\end{equation}
The feasible choice $t=4c_\lambda$ and Taylor's inequality give
\begin{equation}
  R(4c_\lambda P_W)\le R(0)-4kc_\lambda^2+\frac{k}{8}(4c_\lambda)^2=R(0)-2kc_\lambda^2 .
\end{equation}
Since $B=0$ is the only invariant member, \eqref{eq:bern-gap} follows.
\end{proof}

\subsection{The price in bits}

The approximation theorem becomes a compression result only after the code is declared. Consider the idealised two-part population criterion
\begin{equation}
  C_n(\mathcal{A})=K(\mathcal{A})+nR^\star_{\mathcal{A}},\qquad R^\star_{\mathcal{A}}=\inf_{f\in\mathcal{A}}R_P(f),
\end{equation}
and suppose the invariant subclass saves $B=K(\F_k)-K(\F_k^G)>0$ nats in model description.

\begin{proposition}[Exact two-part crossover]\label{prop:crossover}
For any class with gap $\delta_G$,
\begin{equation}
  C_n(\F_k^G)-C_n(\F_k)=-B+n\delta_G .
\end{equation}
When $\delta_G>0$, the non-invariant class is preferred exactly when $n>B/\delta_G$. Under \cref{thm:main},
\begin{equation}
  n_\star=\frac{2\sigma^2B}{U_k-I_k}
\end{equation}
whenever $U_k>I_k$.
\end{proposition}

\begin{proof}
Subtract the two criteria and use $R^\star_{\F_k^G}-R^\star_{\F_k}=\delta_G$; then substitute \eqref{eq:main}.
\end{proof}

The identity is exact for the stated criterion; it is not a universal formula for every MDL protocol~\cite{kotlowski2011maximum}. For sequential prediction, let $A_G$ and $A$ denote fixed learning algorithms for the invariant and unrestricted families. Let $r_t^G,r_t$ be their expected one-step log losses and define the expected prequential codelength and algorithm-description saving by
\begin{equation}
  \mathrm{CL}_n(A)=K(A)+\sum_{t=1}^{n}r_t(A),\qquad B_A=K(A)-K(A_G).
\end{equation}

\begin{proposition}[Prequential learning-curve decomposition]\label{prop:prequential}
\begin{equation}
  \mathrm{CL}_n(A_G)-\mathrm{CL}_n(A)=-B_A+\sum_{t=1}^{n}\left(r_t^G-r_t\right).
\end{equation}
If $\frac1n\sum_{t=1}^{n}\left[(r_t^G-r_t)-\delta_G\right]\to0$, then
\begin{equation}
  \mathrm{CL}_n(A_G)-\mathrm{CL}_n(A)=-B_A+n\delta_G+o(n).
\end{equation}
\end{proposition}

\begin{proof}
Subtract the definitions; the second statement follows by writing the sum as $n\delta_G$ plus the averaged deviation.
\end{proof}

This separates a persistent pseudo-true risk gap from transient estimation and protocol redundancy.

\subsection{Distillation into a student of limited rank}

This subsection states precisely the re-breaking result described in \cref{sec:repair}. Fix the irreducible setting of \cref{cor:onefamily}, choose a rank-$k$ projector $P$, and let the teacher be the orbit average
\begin{equation}\label{eq:teacher}
  T_P(\,\cdot\mid X)=\int_{O(d)}N\!\left(\inner{\lambda gPg^\top}{X}_F,\ \sigma^2\right)d\mu(g),
\end{equation}
an invariant Gaussian mixture with conditional mean $\inner{cI_d}{X}_F$, $c=\lambda k/d$. Distil it into fixed-variance students $\mathcal{S}_\ell=\{N(\inner{M}{X}_F,\tau^2):\ M\succeq0,\ \rank(M)\le\ell\}$ by minimising $D_T(s)=\E_X\,\mathrm{CE}(T(\cdot\mid X),s(\cdot\mid X))$, and let $\delta_G^{\mathrm{KD}}$ be the price of a symmetric student, defined as in \eqref{eq:gap}.

\begin{theorem}[Exact re-breaking under invariant-teacher distillation]\label{thm:distill}
For $0<\ell<d$, every global minimiser of $D_{T_P}$ over $\mathcal{S}_\ell$ has coefficient $M_W=cP_W$ for some $\ell$-dimensional subspace $W$, and is non-invariant. The only invariant feasible coefficient is zero, and
\begin{equation}
  \delta_G^{\mathrm{KD}}(\mathcal{S}_\ell;T_P)=\frac{\ell\lambda^2k^2}{2\tau^2d^2}.
\end{equation}
For equal base and student ranks $\ell=k$, this becomes $\lambda^2k^3/(2\tau^2d^2)$.
\end{theorem}

\begin{proof}
The orbit-closure argument of \cref{thm:closure} applies to soft labels, so an orbit-closed student class has zero distillation gap; the multiplet model gives the quantitative converse. Schur averaging gives the teacher's exact conditional mean $\E_{T_P}[Y\mid X]=\inner{cI_d}{X}_F$ with $c=\lambda k/d$. By conditional bias--variance decomposition,
\begin{equation}
  D_{T_P}(q_M)=C_T+\frac{1}{2\tau^2}\E_X\!\left[\inner{M-cI_d}{X}_F^2\right]=C_T+\frac{\|M-cI_d\|_F^2}{2\tau^2},
\end{equation}
where $C_T$ contains the teacher's conditional variance and does not depend on $M$. \Cref{thm:main} with one irreducible block, target scale $c$, and budget $\ell$ completes the proof.
\end{proof}

\subsection{Several symmetries: spectral capacity curvature}

Constraint interaction has no universal sign law, but in the structured-rank Gaussian model it has an exact capacity interpretation. Let $H_1,H_2\le G$, write $H_{12}=\langle H_1,H_2\rangle$, and retain the exact-symmetry assumptions of \cref{sec:construction}. For any subgroup $H$, define the captured signal energy
\begin{equation}
  V_k(H)=\|M_\star\|_F^2-\min_{\substack{M\succeq0,\ \rank(M)\le k\\ M\rho(h)=\rho(h)M\ \forall h\in H}}\|M-M_\star\|_F^2 .
\end{equation}
Define
\begin{equation}
  F_k(H)=\inf_{q_M\in\F_k^H}R_P(q_M),\qquad \kappa_k=F_k(H_1)+F_k(H_2)-F_k(H_{12})-F_k(e).
\end{equation}
Writing $\Delta_2=F_k(H_2)-F_k(e)$ and $\Delta_{2\mid1}=F_k(H_{12})-F_k(H_1)$ gives $\Delta_{2\mid1}-\Delta_2=-\kappa_k$. Thus $\kappa_k>0$ denotes substitution and $\kappa_k<0$ complementarity. Noncommutativity alone does not determine the sign.

\begin{proposition}[Spectral capacity curvature]\label{prop:curvature}
For the Gaussian model,
\begin{equation}\label{eq:curvature}
  \kappa_k=\frac{V_k(e)+V_k(H_{12})-V_k(H_1)-V_k(H_2)}{2\sigma^2}.
\end{equation}
Thus interaction is a discrete second difference of a capacity-allocation value function. Its sign depends on representation overlap, block sizes, signal spectrum, and budget.
\end{proposition}

\begin{proof}
Equation \eqref{eq:frob} and the definition of $V_k(H)$ give
\begin{equation}
  F_k(H)=R_P(q_{M_\star})+\frac{\|M_\star\|_F^2-V_k(H)}{2\sigma^2}.
\end{equation}
Substitution into the definition of $\kappa_k$ proves \eqref{eq:curvature}.
\end{proof}

\begin{example}[Two exact reflections become complementary at rank one]\label{ex:reflections}
Let $d=2$, $M_\star=\lambda I_2$, and $k=1$. Let $r_1$ and $r_2$ be reflections about lines separated by angle $\pi/n$, with $n\ge3$, and let $H_i=\langle r_i\rangle$. The two reflections do not commute and generate the finite dihedral group $D_n$. Both are exact symmetries of the joint law and target. An optimal rank-one projector can align with an eigenspace of either reflection, so
\begin{equation}
  V_1(H_1)=V_1(H_2)=V_1(e)=\lambda^2 .
\end{equation}
The symmetric commutant of $D_n$ consists of scalar matrices. At rank one its only positive semidefinite member is zero, so $V_1(D_n)=0$. Consequently
\begin{equation}
  \kappa_1=-\frac{\lambda^2}{2\sigma^2}<0,\qquad \Delta_2=0,\qquad \Delta_{2\mid1}=\frac{\lambda^2}{2\sigma^2}.
\end{equation}
Each reflection is free in isolation, yet after choosing a rank-one axis compatible with the first, exact enforcement of the second requires discarding the remaining signal.
\end{example}

The example exhibits strict symmetry interaction without implying that every pair of noncommuting symmetries has negative curvature. The universal statement is the value-function identity \eqref{eq:curvature}; the strict sign is a property of a specified representation, spectrum, and budget.

%% file: appB_attention.tex
\section{Proofs for attention heads}\label{app:attention}

Throughout, $G$ is a compact group acting continuously on $\X$, $\mu$ is a $G$-invariant probability measure, and $L^2(\mu)$ is the real Hilbert space of square-integrable functions on $\X$. A kernel $s\in L^2(\mu\otimes\mu)$ defines the integral operator $(T_sf)(x)=\int s(x,y)f(y)\,d\mu(y)$, which is Hilbert--Schmidt with $\|T_s\|_{\mathrm{HS}}=\|s\|_{L^2(\mu\otimes\mu)}$. Because $\mu$ is $G$-invariant, $(\pi(g)f)(x)=f(g^{-1}x)$ defines an orthogonal representation of $G$ on $L^2(\mu)$; we assume, as usual, that it is strongly continuous.

\begin{proof}[Proof of \cref{lem:rank}]
If $s(x,y)=\sum_{t\le k}\phi_{Q,t}(x)\phi_{K,t}(y)$, then $T_sf=\sum_{t\le k}\inner{\phi_{K,t}}{f}\,\phi_{Q,t}$, whose range lies in the span of $\phi_{Q,1},\dots,\phi_{Q,k}$. On a finite set of queries $x_1,\dots,x_n$ and keys $y_1,\dots,y_n$, the matrix $[s(x_i,y_j)]$ is the product of the $n\times k$ matrices $[\phi_{Q,t}(x_i)]$ and $[\phi_{K,t}(y_j)]^\top$, so its rank is at most $k$. Conversely, an operator of rank at most $k$ can be written $Tf=\sum_{t\le k}\inner{b_t}{f}\,a_t$ with $a_t,b_t\in L^2(\mu)$, and its kernel $\sum_ta_t(x)b_t(y)$ is the logit of the head with $\phi_Q=(a_1,\dots,a_k)$ and $\phi_K=(b_1,\dots,b_k)$.
\end{proof}

\begin{proof}[Proof of \cref{thm:attention}]
\emph{Step 1: the loss is a distance between operators.} A head predicts $N(s(x,y),\sigma^2)$ for $Y=\kappa(x,y)+\varepsilon$, so its population log loss exceeds that of the target similarity by
\begin{equation}
  \frac{\E\big[(s(x,y)-\kappa(x,y))^2\big]}{2\sigma^2}=\frac{\|T_s-T_\kappa\|_{\mathrm{HS}}^2}{2\sigma^2}.
\end{equation}

\emph{Step 2: the best head.} Since $\kappa$ is symmetric and square-integrable, $T_\kappa$ is a compact self-adjoint operator, and by assumption $T_\kappa=\sum_a\lambda_aP_a$ with $P_a$ the orthogonal projection onto the finite-dimensional subspace $V_a$. Its singular values are the $|\lambda_a|$, each repeated $d_a$ times, and $\|T_\kappa\|_{\mathrm{HS}}^2=S=\sum_ad_a\lambda_a^2$. By \cref{lem:rank} the heads are exactly the operators of rank at most $k$, and the Eckart--Young--Mirsky theorem for Hilbert--Schmidt operators~\cite{eckart1936approximation,mirsky1960symmetric} gives
\begin{equation}
  \min_{s\in\Hd_k}\|T_s-T_\kappa\|_{\mathrm{HS}}^2=S-U_k .
\end{equation}

\emph{Step 3: symmetric heads commute with the symmetry.} For any $g$, the kernel of $\pi(g)^{-1}T_s\pi(g)$ is $s(gx,gy)$, by the change of variables $y\mapsto gy$ and invariance of $\mu$. Hence $s$ is symmetric exactly when $T_s$ commutes with every $\pi(g)$.

\emph{Step 4: the best symmetric head.} Because $G$ is compact, $L^2(\mu)$ decomposes into isotypic components, one for each type of irreducible representation. Operators commuting with $\pi$ map each component into itself. Let $H_a$ be the component of the type of $V_a$. Because the $V_a$ are pairwise inequivalent, $V_a$ is the only one of them inside $H_a$, so $T_\kappa$ acts on $H_a$ as $\lambda_aP_a$ and vanishes on every component that contains no $V_a$. Write $H_a\cong W_a\otimes C_a$, with $W_a$ the irreducible representation and $C_a$ its multiplicity space, and let $e_a\in C_a$ be the unit vector with $V_a=W_a\otimes e_a$. Since $W_a$ is absolutely irreducible, an operator commuting with $\pi$ acts on $H_a$ as $I_{W_a}\otimes B_a$ for an operator $B_a$ on $C_a$, with
\begin{equation}
  \rank(I_{W_a}\otimes B_a)=d_a\rank(B_a),\qquad \|I_{W_a}\otimes B_a-\lambda_aI_{W_a}\otimes e_ae_a^\top\|_{\mathrm{HS}}^2=d_a\|B_a-\lambda_ae_ae_a^\top\|_{\mathrm{HS}}^2 .
\end{equation}
If $B_a=0$, block $a$ contributes error $d_a\lambda_a^2$. If $B_a\ne0$, it costs at least $d_a$ units of rank, and its contribution is at least zero, with equality at $B_a=\lambda_ae_ae_a^\top$. On components where $T_\kappa$ vanishes, any nonzero part of $T_s$ only adds error and rank. Hence a symmetric head of rank at most $k$ has error at least $S-\sum_{a\in J}d_a\lambda_a^2$ for some set $J$ of blocks with $\sum_{a\in J}d_a\le k$, and the head $\sum_{a\in J}\lambda_aP_a$ attains it. Minimising over $J$ gives
\begin{equation}
  \min_{s\in\Hd_k^G}\|T_s-T_\kappa\|_{\mathrm{HS}}^2=S-I_k .
\end{equation}

\emph{Step 5: conclusion.} Subtracting the two minima and dividing by $2\sigma^2$ gives \eqref{eq:attention}. If $U_k>I_k$, every optimal head has strictly lower loss than every symmetric head, so no optimal head is symmetric. Finally, $s(g\,\cdot,g\,\cdot)$ has the same loss as $s$ because $\mu\otimes\mu$ and $\kappa$ are invariant, so the optimal heads are closed under the symmetry.
\end{proof}

\begin{proof}[Proof of \cref{cor:sphere}]
Let $\mathcal{H}_l$ be the space of spherical harmonics of degree $l$, of dimension $2l+1$, and let $Q_l$ be the orthogonal projection onto it. By the Funk--Hecke formula~\cite{atkinson2012spherical}, for $f\in\mathcal{H}_l$ and the uniform probability measure,
\begin{equation}
  \int_{S^2}P_j(x\cdot y)\,f(y)\,d\mu(y)=\frac{\delta_{jl}}{2l+1}\,f(x),
\end{equation}
so $T_\kappa=\sum_{l\le L}\lambda_lQ_l$. The $\mathcal{H}_l$ are absolutely irreducible and pairwise inequivalent representations of $SO(3)$, so \cref{thm:attention} applies with blocks of size $2l+1$.

Because $\lambda_0>\cdots>\lambda_L>0$, the strongest $k=l^2+r$ directions, with $0\le r<2l+1$, are the complete shells $0,\dots,l-1$ and $r$ directions of shell $l$. If $r=0$ this set is a union of whole shells, so $U_k=I_k$, and the optimal head $\sum_{l'<l}\lambda_{l'}Q_{l'}$ is unique and symmetric. If $r>0$, a packing of whole shells either uses fewer than $k$ dimensions, and then captures less than $U_k$ because every direction in the first $(L+1)^2$ carries positive energy, or uses exactly $k$ dimensions without being the strongest $k$, and then captures strictly less because it replaces stronger directions by weaker ones. Hence $I_k<U_k$, and every optimal head is broken.

For $r>0$, let $A$ be an optimal head, $R$ its range, of dimension at most $k$, and $Q_R$ the orthogonal projection onto $R$. The operators $T_\kappa-Q_RT_\kappa$ and $Q_RT_\kappa-A$ are orthogonal in the Hilbert--Schmidt inner product, because the range of the second lies in $R$; so $\|T_\kappa-A\|_{\mathrm{HS}}^2\ge\|T_\kappa-Q_RT_\kappa\|_{\mathrm{HS}}^2=S-\tr(Q_RT_\kappa^2)$, with equality only if $A=Q_RT_\kappa$. In an orthonormal eigenbasis $(e_j)$ of $T_\kappa$ with eigenvalues $\mu_j$, $\tr(Q_RT_\kappa^2)=\sum_j\mu_j^2w_j$ with weights $w_j=\|Q_Re_j\|^2\in[0,1]$ summing to at most $k$. This sum equals $U_k$ only if $w_j=1$ whenever $\mu_j^2>\lambda_l^2$ and $w_j=0$ whenever $\mu_j^2<\lambda_l^2$. Hence $R$ contains the shells $0,\dots,l-1$, lies inside them together with $\mathcal{H}_l$, and meets $\mathcal{H}_l$ in an $r$-dimensional subspace $W$. The optimal heads are therefore exactly $T=F+\lambda_lQ_W$, where $F=\sum_{l'<l}\lambda_{l'}Q_{l'}$ and $Q_W$ projects onto an $r$-dimensional subspace $W$ of $\mathcal{H}_l$. The kernel of $\pi(g)^{-1}T\pi(g)$ is $s(gx,gy)$, so the rotation sensitivity equals
\begin{equation}
  \E_g\big\|T-\pi(g)^{-1}T\pi(g)\big\|_{\mathrm{HS}}^2=2\|T\|_{\mathrm{HS}}^2-2\|\Pi T\|_{\mathrm{HS}}^2,
\end{equation}
where $\Pi$ averages over conjugation by the symmetry and is an orthogonal projection. $F$ is invariant, and by Schur's lemma $\Pi Q_W=\frac{r}{2l+1}Q_l$. Since $F$ and $Q_W$ act on different shells,
\begin{equation}
  \|T\|_{\mathrm{HS}}^2-\|\Pi T\|_{\mathrm{HS}}^2=\lambda_l^2r-\lambda_l^2\frac{r^2}{2l+1},
\end{equation}
which gives \eqref{eq:sphere-dispersion}.
\end{proof}

\paragraph{Softmax heads.} For completeness, the claim of \cref{sec:softmax-theory} that symmetric softmax heads use whole blocks follows from the same argument. Centring a head's logits over keys replaces $T_s$ by $T_s(I-Q_0)$, which still has rank at most $k$. If the centred logits are invariant, $T_s(I-Q_0)$ commutes with the symmetry, and Step 4 shows that it acts on each shell $l\ge1$ as a multiple of the identity.

%% file: appC_experiments.tex
\section{Experimental details}\label{app:experiments}

\paragraph{Settings.} \Cref{tab:settings} lists the configuration of the GPU runs reported in \cref{sec:experiments}.

\begin{table}[h]
\centering\small
\caption{Configuration of the GPU runs.}\label{tab:settings}
\begin{tabular}{@{}ll@{}}
\toprule
Tokens & independent uniform unit vectors in $\R^3$\\
Target similarity & \eqref{eq:sphere-target} with $\lambda=(1,0.8,0.6,0.45,0.33,0.25)$, shells $l=0,\dots,5$\\
Head & separate query and key MLPs, SiLU, $3\to128\to128\to128\to k$; logit $=$ dot product\\
Initialisation & weights $N(0,2/\text{fan-in})$, zero biases\\
Logit regression & squared error on 8,192 fresh query--key pairs per step\\
Softmax attention & KL to $\mathrm{softmax}(m\kappa)$ over 64 keys, 256 fresh queries per step\\
Optimiser & Adam, learning rate $2\times10^{-3}$, 100 warm-up steps, cosine decay to $10^{-5}$\\
Main sweep & both objectives, $k=1,\dots,20$, $m=0.3$, 3 seeds, 1,500 steps (120 models)\\
Temperature follow-up & softmax, $k=6,\dots,10$, $m\in\{0.1,0.3\}$, 3 seeds, 6,000 steps (30 models)\\
Evaluation & 100,000 pairs (logit regression); 4,096 query sets of 64 keys (softmax)\\
Hardware and software & NVIDIA B200 GPUs; JAX 0.12.0.1; CUDA 12 (main sweep), CUDA 13 (follow-up)\\
\bottomrule
\end{tabular}
\end{table}

\paragraph{Rotation sensitivity.} For logit regression, each evaluation pair $(x,y)$ receives an independent uniformly random rotation $g$, and we report the mean of $(s(x,y)-s(gx,gy))^2$. For softmax attention, each query and its 64 keys receive one common rotation; logits are centred by subtracting each query's mean over its keys, and we report the mean squared change of the centred logits. In \cref{fig:softmax} and \cref{tab:softmax,tab:spread}, values are divided by $m^2$ so that the two target scales can be compared.

\paragraph{Protocol.} The predicted symmetric head dimensions ($k=1,4,9,16$ for logit regression and $k=3,8,15$ for softmax attention) and the predicted sensitivities \eqref{eq:sphere-dispersion} were written to file before any training. The softmax prediction failed at $k=8$ under peaked attention. The temperature follow-up was declared after that result; its hypothesis, that flatter attention would restore the symmetric head at $k=8$, was recorded before the follow-up runs. Every trained model is reported, the final model of each run is evaluated, and no selection based on rotation sensitivity occurs. Error bars are standard deviations over seeds, not confidence intervals.

\paragraph{Software versions.} The main sweep ran with the CUDA 12 build of JAX, which issued a cuBLAS warning on these GPUs. We therefore repeated 69 of its 120 models with the CUDA 13 build: all 60 logit-regression models, and the softmax models at $k=3,8,15$. The repeats reproduced the measured sensitivities to within $1.2\times10^{-7}$. The remaining 51 softmax models of the main sweep come from the CUDA 12 build only. The temperature follow-up used the CUDA 13 build.

\paragraph{Best whole-shell heads.} For a set $J$ of shells, the symmetric head with logits $\sum_{l\in J}c_l(2l+1)P_l(x\cdot y)$ uses $\sum_{l\in J}(2l+1)$ dimensions; the constant shell is omitted because the softmax ignores it. We fitted every set $J\subseteq\{1,\dots,5\}$ with at most 20 dimensions (19 sets) at both multipliers. The coefficients start at $c_l=m\lambda_l$ and are fitted by Adam (learning rate $5\times10^{-3}$) for 2,000 steps on fresh batches of 1,024 queries with 64 keys; the KL is evaluated on $8\times4{,}096$ held-out query sets. The best whole-shell head of dimension $k$ is the best over all $J$ that fit within $k$; for $k<3$ it is the constant head, which gives uniform attention. Fits made with the exact-feature script, on different evaluation sets, agree within 0.002.

\paragraph{Exact harmonic features.} For each degree $l\le5$ we built an orthonormal basis of $\mathcal{H}_l$ from the functions $P_l(x\cdot a_i)$ for $2l+1$ well-conditioned random directions $a_i$, orthonormalised with their exact Gram matrix $P_l(a_i\cdot a_j)/(2l+1)$, which follows from the Funk--Hecke formula. Stacking the bases gives 36 features $Y(x)$; the largest deviation of their empirical second-moment matrix from the identity on 200,000 samples was 0.007. A head with exact features has logits $Y(x)^\top AB^\top Y(y)$ with $A,B\in\R^{36\times k}$, initialised with independent $N(0,0.05^2)$ entries and trained by Adam for 4,000 steps on 256 queries with 64 keys per step. Further runs started from initial standard deviation 0.3 (three seeds each at $k=8$ and $10$), and at $k=10$ from the symmetric head built from shells 1 and 3 plus Gaussian noise of standard deviation 0.001 (three seeds, learning rate $5\times10^{-4}$), 0.05 or 0.15 (two seeds each). These diagnostic runs, and the MLP runs in \cref{tab:spread}, were made with JAX on CPU and used 1,500 (MLP) or 4,000 (exact features) steps.

\paragraph{Shell use.} For a head we form its $36\times36$ harmonic coefficient matrix: $AB^\top$ for exact-feature heads, and, for MLP heads,
\[
  \E[Y(x)\phi_Q(x)^\top]\,\E[Y(y)\phi_K(y)^\top]^\top,
\]
estimated on 100,000 samples. We keep the right singular vectors whose singular values exceed 1\% of the largest, and report, for each shell, the sum of their squared components in that shell. A head that uses a whole shell scores its full dimension $2l+1$. For MLP heads we also report the fraction of the variance of the key features that lies in each shell.

\paragraph{Convergence of the softmax runs.} For softmax attention at $m=0.3$ and $k=6,\dots,10$, the 6,000-step runs changed the mean KL of the 1,500-step runs by at most 0.005 (0.006 for a single seed) and left every head at least as broken.

\paragraph{Which shells the heads use.} \Cref{tab:spread} lists the shell use behind \cref{sec:spread}. Starting the exact-feature heads from larger random weights (standard deviation 0.3 instead of 0.05) gives heads that spread in the same way and end at higher KL: 0.24--0.25 at $k=8$, still below the best whole-shell head, and 0.18 at $k=10$. Started near the symmetric head built from shells 1 and 3, with noise of standard deviation up to 0.15 in every weight, exact-feature heads stay close to it; trained from small random weights they use about 2.7 dimensions of shell 2, which that head leaves empty. Even at $m=0.1$ the values of shells are not additive: shells 1 and 3 together beat shells 1 and 2 (KL 0.0222 against 0.0228), although shells 1 and 2 carry more energy (3.72 against 3.34).

\begin{table}[h]
\centering\small
\caption{Which shells softmax heads use ($m=0.3$). \emph{Dimensions used} counts how many of the $3,5,7,9,11$ dimensions of shells $l=1,\dots,5$ the head's key features span. \emph{Exact features} replace the multilayer perceptrons with spherical harmonics up to degree five: three random starts at $k=8,10$ and two at $k=15$, with initial weights of standard deviation 0.05, and at $k=10$ seven starts near the symmetric head built from shells 1 and 3. The MLP rows are single CPU runs of the network of the GPU sweep; the sweep itself, with three seeds, gives KL 0.156 and 0.091 and scaled sensitivity 11.9 and 0.21 at $k=10$ and $15$.}\label{tab:spread}
\setlength{\tabcolsep}{3.5pt}
\begin{tabular}{@{}clcccc@{}}
\toprule
$k$ & Head features & KL & Sensitivity$/m^2$ & Dimensions used, $l=1,\dots,5$ & Best whole-shell\\
\midrule
8  & exact features         & 0.208--0.212 & 15.1--15.6 & 2.3 / 2.9 / 2.3 / 0.4 / 0.1 & 0.259\\
10 & MLP                    & 0.155        & 11.3       & 1.8 / 4.0 / 3.1 / 0.5 / 0.1 & 0.138\\
10 & exact features         & 0.150--0.153 & 12.3--13.4 & 2.5 / 2.7 / 3.5 / 1.0 / 0.2 & 0.138\\
10 & exact, symmetric start & 0.137--0.138 & 0.001--0.92 & 2.7--3.0 / $\le$0.1 / 6.6--7.0 / $\le$0.1 / 0.0 & 0.138\\
15 & MLP                    & 0.091        & 0.15       & 2.9 / 4.8 / 6.4 / 0.0 / 0.0 & 0.091\\
15 & exact features         & 0.078--0.081 & 8.5--8.6   & 2.7 / 3.2 / 4.1 / 4.1 / 0.8 & 0.091\\
\bottomrule
\end{tabular}
\end{table}

\paragraph{Orientation of broken heads.} When the head dimension cuts the shell $l=1$ ($k=2,3$ in logit regression, $k=1,2$ under softmax attention), each head singles out an axis in space. The mean absolute cosine between the axes of different seeds is 0.85 at $k=2$ and 0.19 at $k=3$, and 0.46 and 0.36 under softmax attention at $k=1$ and $2$; uniformly random axes give 0.5 on average. With three seeds per setting this test has little power. The seeds at $k=2$ chose similar axes; nothing in the task prefers an axis, but Adam, which rescales each weight separately, is not rotation-equivariant.

\paragraph{Reference implementation checks.} \Cref{tab:checks} reports deterministic, quadrature and Monte Carlo checks of the reference implementation against the analytical formulas of \cref{sec:theorem,app:theory}. They validate the implementation and are not empirical evidence for the population theorems. The checks use seed 20260822, 300,000 symmetric-Gaussian input draws, 40,000 Haar orthogonal matrices, and 180,000 full Gaussian matrices for the Bernoulli result; the Bernoulli reference uses deterministic quadrature and one-dimensional golden-section minimisation, and the remaining discrepancies are Monte Carlo error. The reference run used Python 3.14.3 and NumPy 2.4.2 on one Apple M4 Pro CPU, took 0.69 seconds and used 174,211,072 bytes of resident memory.

\begin{table}[h]
\centering\small
\caption{Checks of the reference implementation.}\label{tab:checks}
\begin{tabular}{@{}lll@{}}
\toprule
Check & Theory / reference value & Numerical result\\
\midrule
Equal-block gap, $r=3$, $k=5$, $\sigma=0.9$ & 0.79012346 & 0.79012346\\
Repeated-isotypic gap, $k=4$, $\sigma=1.2$ & 0.31250000 & 0.31250000\\
Irreducible gap, $d=5$, $k=2$, $\lambda=0.7$ & 0.49000000 & 0.49021011\\
Invariant anisotropy, $\varepsilon=0.12$ & $[0.254800,\ 0.725200]$ & 0.61465600\\
Haar mean $\|\widehat{\Pi M}-\Pi M\|_F$ & 0 & 0.00463236\\
Rank before / exact Haar mean & 2 / 5 & 2 / 5\\
Orbit defect & 1.17600000 & 1.18109443\\
Distillation gap, $\ell=3$, $\tau=1.3$ & 0.06958580 & 0.06964882\\
Bernoulli gap, $d=5$, $k=2$, $\lambda=0.6$, $M=6c_\lambda\sqrt{k}$ & 0.05236399 & 0.05237311\\
Bernoulli lower bound & $\ge0.04970757$ & passed\\
Joint reflection-commutant dimension & 1 & 1\\
\bottomrule
\end{tabular}
\end{table}

\paragraph{Reproducing the runs.} With JAX installed for the GPU, the three groups of GPU runs are
\begingroup\footnotesize
\begin{verbatim}
python attn_sawtooth.py --dh 1-20 --seeds 3 --steps 1500 --out initial
python attn_sawtooth.py --variants B --dh 6-10 --temp 0.1 --seeds 3 \
       --steps 6000 --out flat
python attn_sawtooth.py --variants B --dh 6-10 --temp 0.3 --seeds 3 \
       --steps 6000 --out peaked
\end{verbatim}
\endgroup
{\sloppy\noindent The CPU diagnostics behind the best whole-shell heads and \cref{tab:spread} are supplied with their logs as four scripts: \texttt{whole\_shell\_fits.py}, \texttt{exact\_head.py}, \texttt{exact\_head2.py} and \texttt{mlp\_occ.py}. Each takes 4 to 18 minutes on two CPU cores.\par}

%% file: appD_practice.tex
\section{The procedure in practice}\label{app:practice}

\paragraph{Blocks of common symmetries.} \Cref{tab:blocks} lists the blocks of some symmetries that arise in practice. Under squared error, or with flat attention, a symmetric model can be optimal only at budgets where the strongest directions form whole blocks. Under peaked softmax attention the ranking of blocks can change (\cref{sec:spread}).

\begin{table}[h]
\centering\small
\caption{Blocks of some common symmetries. The budgets listed assume that block strengths decrease in the order given. Under softmax attention the constant block is free (\cref{sec:softmax-theory}).}\label{tab:blocks}
\begin{tabular}{@{}llll@{}}
\toprule
Symmetry of the items compared & Block sizes & Whole-block budgets & Under softmax\\
\midrule
Rotations and reflections of angles & $1,2,2,2,\dots$ & $1,3,5,\dots$ & $2,4,6,\dots$\\
Rotations of directions in $\R^3$ & $1,3,5,7,\dots$ & $1,4,9,16,\dots$ & $3,8,15,\dots$\\
Rotations of directions in $\R^d$, $d\ge3$ & $1,\ d,\ \tfrac{d(d+1)}{2}-1,\dots$ & $1,\ d+1,\ \tfrac{d(d+3)}{2},\dots$ & $d,\ \tfrac{d(d+3)}{2}-1,\dots$\\
Relabelling $n$ equally likely symbols & $1,\ n-1$ & $1,\ n$ & $n-1$\\
\bottomrule
\end{tabular}
\end{table}

\paragraph{Reading the outcome.} \Cref{tab:outcomes} summarises how to read steps 4 to 7 of the procedure in \cref{sec:method}. ``High'' sensitivity means well above the values at budgets where whole blocks fit. To compare different scales, divide the rotation sensitivity by twice the mean squared (centred) logit, which gives a number between 0 (symmetric) and 1 (no symmetric component).

\begin{table}[h]
\centering\small
\caption{Reading the outcome of the procedure. Losses are compared at the same budget.}\label{tab:outcomes}
\begin{tabular}{@{}>{\raggedright\arraybackslash}p{3.1cm}>{\raggedright\arraybackslash}p{2.2cm}>{\raggedright\arraybackslash}p{5.6cm}>{\raggedright\arraybackslash}p{4.6cm}@{}}
\toprule
Trained loss versus best symmetric loss & Sensitivity & Reading & Action\\
\midrule
lower & high & breaking is the price of capacity & complete the blocks, add a cheap invariant path, or accept it\\
equal & near zero & a symmetric solution was reached; if strong blocks are left unused, the features may be the limit (\cref{sec:spread}, $k=15$) & none, or richer features\\
equal & high & breaking neither helps nor hurts at this budget (\cref{sec:softmax-results}, $k=11$) & none needed; a symmetric solution is available\\
higher & high & not the price of capacity: training ended at a worse broken solution (\cref{sec:spread}, $k=10$) & start from, or constrain to, the symmetric solution\\
any & high, and runs agree on the orientation & suspect the data, architecture or optimiser & check the data and the symmetry action\\
\bottomrule
\end{tabular}
\end{table}

\paragraph{Remedies and their costs.} If the violation is the price of capacity, there are four options.
\begin{itemize}
  \item \textbf{Complete the blocks.} Raise the budget to a value where the strongest directions form whole blocks ($U_k=I_k$ in \cref{thm:attention}). Under squared error this removes the price. Under peaked softmax attention it did not: the head at $k=8$, where shells 1 and 2 fit exactly, still broke the symmetry.
  \item \textbf{Make the symmetric answer cheap.} Give the model a path that computes invariants directly, such as distances or angles between tokens, or use an architecture whose interactions are equivariant by construction~\citep{cohen2016group,thomas2018tensor,fuchs2020se3}. This changes the class, so the budget problem disappears.
  \item \textbf{Average, then distil with room to spare.} Averaging over the symmetry restores it exactly and never increases the loss: for averaged predictions by \cref{thm:closure}, and for averaged logits because squared error and the softmax KL are convex in the logits. But the averaged model uses the whole of every block the broken model touched. A softmax head that uses part of shells 1 to 5 averages to a head that needs all 35 of their dimensions (\cref{tab:spread}). A student without room for the blocks it needs can break the symmetry again, as \cref{thm:distill} proves for a single block.
  \item \textbf{Leave it.} The broken model has the lower loss. A repair is worth its cost when consistent outputs on equivalent inputs, or agreement between training runs, matter more than the loss.
\end{itemize}

\paragraph{How large is the price?} If the budget affords $r$ directions of a block whose directions each carry energy $\lambda^2$, the price in \cref{thm:attention} is at most $r\lambda^2/(2\sigma^2)$. It is large when the cut block carries much of the signal and small when the cut falls in the weak tail of the spectrum. For directions in three dimensions, a head of dimension 64 or 128 has room for all shells up to degree 7 or 10, so any cut falls at degree 8 to 11, where a smooth similarity carries little signal. Small heads, or sharp similarities whose signal sits in high degrees, put strong blocks at the cut. Blocks also grow quickly with the input dimension: for directions in $\R^{16}$, the first blocks after the constant have 16, 135 and 800 dimensions, so a symmetric head of dimension 64 can use the linear block, which gives the cosine similarity, but no part of the quadratic one.

\paragraph{An example with one large block.} Suppose a head must score whether two tokens are identical, and all $n$ tokens of a vocabulary are equally likely. Relabelling the vocabulary is then a symmetry, with blocks of sizes 1 and $n-1$. A symmetric head of dimension below $n-1$ can only give every pair of tokens the same score, so any useful head must break the symmetry by representing some tokens' identities better than others, and which tokens it favours is arbitrary. Real vocabularies are far from uniform, which removes this exact symmetry, but the example shows the scale involved: a block of size $n-1\approx50{,}000$ against head dimensions of 64 to 128.

\paragraph{The symmetry must act on the pair.} The bottleneck matters only when the target score depends on how the two items relate, such as the angle between two directions, and the symmetry moves both items together. If the score is invariant for each item separately, for example because it depends only on the lengths of two vectors, each side can compute its invariants alone and no budget is spent. Molecular and point-cloud models that already feed distances or angles into attention have such a path for the relations those features cover.

\paragraph{Depth and peakedness.} In a deep model, earlier layers can move one token's features into another token's position, after which a per-token network can compare them without spending the head's budget; the result concerns the head that performs a comparison, not a model's total capacity. In our task, flatter target attention (perplexity about 60 of 64 keys) restored the predicted symmetric head, while peaked attention (perplexity about 24) broke the symmetry more widely. This suggests that a head's attention perplexity indicates which regime applies, though we have tested only two values.